# BERT Fine-Tuning for Sentiment Analysis on Indonesian Mobile Apps Reviews

*Kuncahyo Setyo Nugroho*

Faculty of Computer Science, Brawijaya University, Indonesia, ksnugroho26@gmail.com

*Anantha Yullian Sukmadewa*

Faculty of Computer Science, Brawijaya University, Indonesia, ananthayullian@gmail.com

*Haftittah Wuswilahaken DW*

Faculty of Computer Science, Brawijaya University, Indonesia, haftittah@gmail.com

*Fitra Abdurrachman Bachtiar*

Faculty of Computer Science, Brawijaya University, Indonesia, fitra.bachtiar@ub.ac.id

*Novanto Yudistira*

Faculty of Computer Science, Brawijaya University, Indonesia, yudistira@ub.ac.id

User reviews have an essential role in the success of the developed mobile apps. User reviews in the textual form are unstructured data, creating a very high complexity when processed for sentiment analysis. Previous approaches that have been used often ignore the context of reviews. In addition, the relatively small data makes the model overfitting. A new approach, BERT, has been introduced as a transfer learning model with a pre-trained model that has previously been trained to have a better context representation. This study examines the effectiveness of fine-tuning BERT for sentiment analysis using two different pre-trained models. Besides the multilingual pre-trained model, we use the pre-trained model that only has been trained in Indonesian. The dataset used is Indonesian user reviews of the ten best apps in 2020 in Google Play sites. We also perform hyper-parameter tuning to find the optimum trained model. Two training data labeling approaches were also tested to determine the effectiveness of the model, which is score-based and lexicon-based. The experimental results show that pre-trained models trained in Indonesian have better average accuracy on lexicon-based data. The pre-trained Indonesian model highest accuracy is 84%, with 25 epochs and a training time of 24 minutes. These results are better than all of the machine learning and multilingual pre-trained models.

## 1 INTRODUCTION

Apps are developed to meet human needs in various aspects of life. Meanwhile, mobile apps are specifically designed to run on mobile phone devices such as smartphones or tablets. The increasing human need has become a business opportunity to develop many mobile apps. Some factors affect users for using apps, such as usability, ratings, and ease of use [1]. In the apps development process, the usability factor is the most necessary aspect because it can reflect the level of ease apps when used [2]. Usability evaluation always involves users, one of them is user reviews. User reviews have rich information when a user is using the apps. Usually, user reviews have two essential parts are rating and opinion. The rating indicates the overall evaluation of the user experience using a numerical scale. While opinion is a deeper story about the user is experienced when using the apps. Ratings do not necessarily provide honest reviews. In addition, ratings do not contain information to increase the user's usability factor. Reading the opinions of user reviews can provide better insight and comprehension. There are no restrictions on how users submit text-based reviews. This makes the review dataset challenging to work with, as it is less structured. Therefore, user opinion creates complexity if read manually by humans. Sentiment analysis can be used to conclude the textual content of each opinion, whether the user likes or dislikes it when using the apps. By knowing user preferences through reviews, developers can increase the usability factor so that they can develop the business even further.

Sentiment analysis, also known as opinion mining classifies user opinion based on polarity. Sentiment analysis consists of different tasks, methods, and types of analysis. Three methods used in sentiment analysis are machine learning (ML), hybrid learning, and lexicon-based [3]. The most popular and widely used method of machine learning is supervised learning. This approach trains a model using labeled data to predict output by considering new unlabeled data [4]. The most used model are naïve bayes (NB) classifier [5], [6] and support vector machines (SVM) [7], [8]. Although this method has shown promising results, this method requires feature engineering to represent the text [9]. The more complete the features used to capture information, the better the classification results, complicating the overall machine learning process. Meanwhile, the traditional method for sentiment analysis is lexicon-based [10]. This method inspects each sentence for words that express a positive or negative feeling to classify sentiment. Because sentiment analysis is domain-dependent, therefore the lexicon used needs to be matched with that domain. It can be a disadvantage of the lexicon-based method, as it is a time-consuming process. However, the lexicon-based method does not require training data, so this is advantageous [11]. There are two main approaches to creating sentiment lexicons: dictionary-based [12], [13] and corpus-based [14], [15]. In many cases, the dictionary-based approach works well for the general domain. Meanwhile, corpus-based lexicons can be adapted to certain domains. The hybrid method is used to extend machine learning models with lexicon-based knowledge [16]. With this, the limitations of both previous approaches can be overcome.

Deep neural networks (DNNs) have recently shown remarkable results and are widely used in various research. Two main DNN architectures are convolutional neural networks (CNNs) and recurrent neural networks (RNNs). Generally, CNNs are hierarchical architectures, while RNNs are sequential architectures [17]. CNN shows the best results in image processing [18], and the same architecture has been applied to text processing [19]. Kim and Jeong used CNN architecture for sentiment analysis on three different datasets [20]. The results show that employing consecutive convolutional layers effectively for longer text and provide better results than various machine learning methods. Meanwhile, Kuniasari and Setyanto used the RNN architecture for sentiment analysis in Indonesian [21]. The results show that RNN has better accuracy than CNN, although the RNN model may overfitting during the testing process. RNN also suffers from vanishing gradients problem, which hinders learning for long text sequences. Long short-term memory (LSTM) enhances the RNN architecture to solve the vanishing gradients problem by taking long-term dependencies. [22]. Therefore, LSTM reportedly outperformed both CNN and RNN for sentiment analysis [23], [24]. More recently, the language model (LM) is a large network trained on unlabeled data and can be fine-tuned to downstream tasks has made breakthroughs in NLP.

One of the language model approach successes is obtained by transferring the encoder from an encoder-decoder architecture based on transformers [25]. Successful language models are GPT-2 [26] by Open AI[1], BERT [27] by Google AI Language Team[2], and XLNet [28] by Google AI Brain Team[3]. BERT is of the best-performing models because it can improve understanding of a context. Therefore, this model has several developments, such as RoBERTa [29] to improving training procedures and ALBERT [30] by reducing the model size. BERT also provides a pre-trained model that is readily and easily fine-tuned across a variety of downstream tasks. Besides that, pre-trained models are also available in various languages [31]–[33]. Research from [34] conducted an in-depth experiment to investigate the effectiveness of the BERT fine-tuning for text classification, show the latest state-of-the-art result in the sentiment analysis of reviews. Nguyen et al. using BERT fine-tuning for sentiment analysis in Vietnamese, show that BERT outperforms other LSTM, TextCNN, RCNN with impressive results [35].

Based on recent studies and the need for user evaluation of mobile apps in Indonesian, we conducted a sentiment analysis using BERT fine-tuning. Although much sentiment analysis research has been conducted on the Indonesian dataset, in-depth experiments using BERT fine-tuning for apps review in Indonesian have never been done before. The main contribution of this paper is to determine the effectiveness of the fine-tuning method for sentiment analysis in Indonesian. We collected 10,615 Indonesian user reviews on the 10 best 2020 apps. We use two different pre-trained models, the multilingual BERT-Base [27] trained on multilingual data and the IndoBERT-Base [31], specially trained on the Indonesian dataset. Next, we perform hyper-parameter tuning for best results. As evaluation criteria, we use accuracy and computation time for each model. The paper is structured as follows: Section 2 describes the proposed methodology while Section 3 describes the experimental design and experimental results for discussion, and Section 4 concludes the work and future work.

## 2 METHODOLOGY

### 2.1 Data collection

Dataset consists of apps reviews in Indonesian on the ten 2020 best-rated apps[4]. We collected 10,615 user reviews using scraping techniques on the Google Play website[5]. Each user review includes the following information as a column: review id, username, user image, content, score, and review date. We only use the content and score columns in this study. The data collected is unlabeled. Meanwhile, to do classification task, labeled training data is required. It is a long and expensive process to label one at a time. Therefore, in this study, we use two labeling approaches to divide the dataset into three classes: positive, neutral, and negative. We provide labeling based on the given user review score in the first approach, as shown in Figure X. The dataset has five score ranges from 1 to 5. A smaller score indicates that the worse the review is given. Reviews scored 1 and 2 are labeled negative, reviews scored 4 and 5 are labeled positive, the rest are labeled neutral. The second approach applies the lexicon method to extract review sentences, as shown in Figure X. We use the InSet Lexicon [36], which contains 3,609 positives and 6,609 negatives in Indonesian words. Corpus identifies the word that contains negative or positive sentiments for each review in the dataset. Then, each word in the sentence is weighted. If the sentence has a positive final score, it is positive class, vice versa. However, if the words in the sentence have no meaning, the sentiment is given a zero value which means it is a neutral class.

---

[1] https://openai.com.
[2] https://research.google/teams/language.
[3] https://research.google/teams/brain.
[4] https://www.appannie.com/en/insights/market-data/2020-mobile-recap-how-to-succeed-in-2021.
[5] https://play.google.com/store.

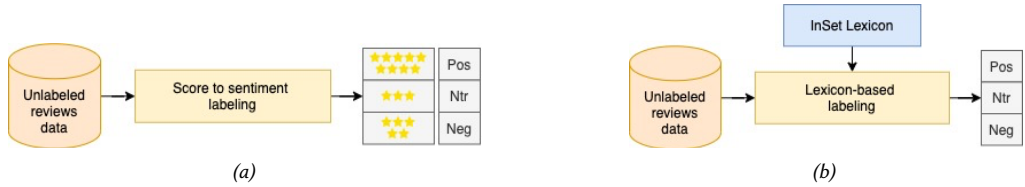

*Figure 1: Training data labeling method to obtain sentiment polarity, (a) using the score to sentiment while (b) using the lexicon-based method with InSet Lexicon*

Table 1: Parameters comparison of pre-trained models. L=numbers of transformer layers, H=numbers of hidden embedding size, and A=numbers of attention heads.

| Pre-Trained Model | L | H | A | Total Parameter | Language Type |
|---|---|---|---|---|---|
| BERT-Base multilingual | 12 | 768 | 12 | 110 million | Multi language (104) |
| IndoBERT-Base | 12 | 768 | 12 | 124.5 million | Mono language |

## 2.2 BERT fine-tuning

For downstream tasks such as sentiment analysis on apps reviews, BERT can be used with two approaches. As feature extraction, model architecture is preserved and outputs are input feature vectors for subsequent classification models. Another method is to modify model architecture by adding some layers at the end. These layers will solve the downstream tasks by retraining the model. As a starting point, models are trained on a huge dataset in the pre-training stage. Then, the model is retrained on a relatively small dataset for a specific domain. The process of retraining model is named fine-tuning. BERT is a giant neural network architecture that has a million parameters. Hence, pre-training a model on a small dataset from scratch would result in an overfitting model. In this study, we fine-tuning BERT uses a pre-trained model to solve sentiment analysis problem on apps reviews. Pre-trained models are providing a context to words that have previously been learning the occurrence and representations of words from unlabeled training data.

A pre-trained model must be trained using the specific language. Our dataset contains user reviews using Indonesian, therefore we need pre-trained models are trained in Indonesian. We use two pre-trained models for fine-tuning. First, BERT-Base Multilingual provided by researchers from Google AI Language who trained in 104 languages, including Indonesian [27]. We also use pre-trained models who are specially trained in Indonesian, named IndoBERT-Base [31]. Parameters comparison of pre-trained models is shown in Table 1. In the original paper, L represents the numbers of transformer layers (stacked encoder), H represents numbers of hidden embedding size, and A represents the numbers of attention heads. Although both pre-trained models have the same number of layers, IndoBERT-Base has a higher number of parameters as in Table 1.

Raw text in the dataset must be adjusted to BERT input format before fine-tuning. Input formatting required a tokenizer by adding a special token for each sentence, as shown in Figure 2. The tokenizer must match that provided by the pre-trained model. This is because the pre-trained model has a specific fixed vocabulary and the tokenizer has a particular way of handling out-of-vocabulary words. For classification, a [CLS] token must be added at the beginning of each sentence. While, at the ending of the sentence, a [SEP] token must be added. Tokenizer has two limitations for handling varying sentence lengths in the dataset. Each sentence must pad or truncate to the maximum sentence length is 512 tokens. A special [PAD] token is used to add padding. The "attention mask" consists of '1' and '0' where '1' indicates the token and '0' is not. This mask tells "self-attention" not to include this [PAD] token when interpreting sentences.

To fine-tuning the model for classification, apply an additional fully-connected dense layer on top of its output layer and retrain the entire model with the specific domain dataset. BERT only takes the final hidden state of the [CLS] token as a representation of the entire sentence to feed into a fully-connected dense layer. The size of the last fully-connected dense layer is equal to the number of labels. SoftMax activation with sparse categorical cross-entropy is added on top of the model to predict the likelihood of the label. Figure 2 shows how fine-tuning model is used for classification. According to the original paper, the author gives a range of values to fine-tuning for the downstream tasks. Batch size range [16, 32], learning rate (Adam) range [5e-5, 3e-5, 2e-5], and number of epochs [2, 3, 4].

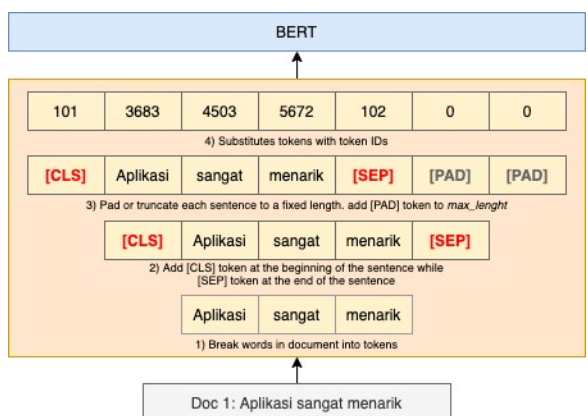

*Figure 2: BERT input formatting schema that requires tokenizer with adding special token*

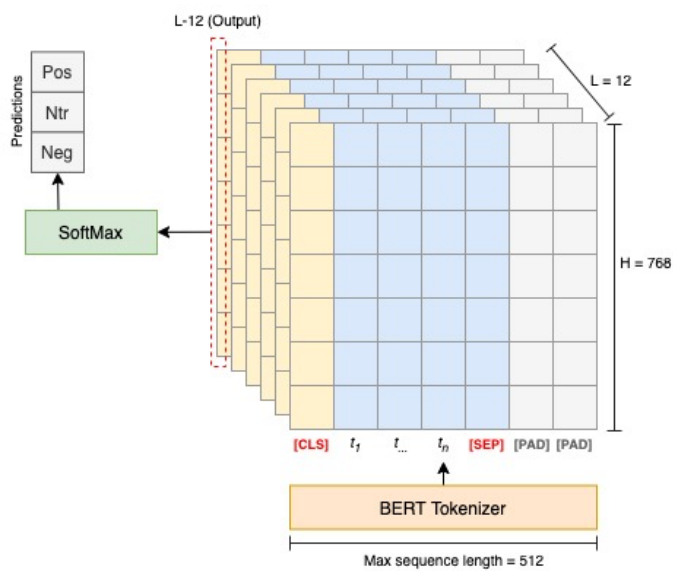

*Figure 3: BERT fine-tuning for sentiment analysis*

## 2.3 Evaluation criteria

We use accuracy to measure the evaluation performance of all models. One of the more obvious metrics, it is the measure of all the correctly identified cases. It is most used when all the classes are equally important. Accuracy is used when the true positives (TP) and true negatives (TN) are more important. Accuracy defined in Equation (1). Where, TP = True Positive, TN = True Negatives, FP = False Positives and FN = False Negatives. We also calculate the computation time during the training phase.

$$accuracy = \frac{TP + TP}{TP + TN + FP + FN} \quad (1)$$

## 3 RESULT AND DISCUSSION

The data that has been collected is unstructured. Therefore, we do basic text pre-processing to clean up the data, including removing numbers, removing unnecessary words, and deleting the name of the application used as keyword scrapping. Stemming is not done because it can change the meaning of the whole sentence. Then, the clean data is labeled using two different methods. Using the score-based method, the results of data labeling show the average distribution of sentiment classes, as shown in Figure 4. While using the lexicon-based method shows the imbalanced distribution of sentiment classes. The number of positive sentiments in the lexicon-based is more than others, as shown in Figure 4. For fine-tuning phase, the data is divided into three parts, training set, test set, and validation set. A total of 9,553 data were used as a training set, while test and validation data have the same 531 data.

All of experiments are run on Google Colab[6] using Python 3.6 which has 1 Tesla V100-SXM2-16GB GPU and 27.8 RAM. For the fine-tuning experiment environment, we use PyTorch[7] version 1.7.0 and transformers[8] version 3.5.1. We use two pre-trained models, BERT-Base multilingual and IndoBERT-Base. We fine-tune each pre-trained model with Adam optimizer for each task with initial learning with a range of learning rates [1e-5, 2e-5, 3e-5], the base learning rate is [1e-5]. We apply a decay rate of [1e-4] for every epoch and sample each batch with a size of 16 and 32 to ensure that the GPU memory is fully utilized. We set the max number of the epoch to [10, 25] and save the best model on the validation set for testing.

Before fine-tuning phase, we conducted an experiment using a machine learning model (ML) as a baseline. The five models we use are k-nearest neighbor (kNN), naïve bayes (NB) classifier, support vector machine (SVM), decision tree, and random forest. The results of machine learning model training are shown in Table 2. At one-fold validation, almost all models provide high accuracy, which shows that the model is overfitting. Overfitting occurs because the model is too focused on the training set, so it cannot predict correctly if a testing set is provided. Therefore, we use ten-fold cross-validation in model training. In the score-based labeling dataset, the highest test accuracy is obtained SVM with 0.5736 accuracy. Whereas in the lexicon-based labeling dataset, the random forest provides the best accuracy in 0.7890. The test results of the baseline experimental show that the lexicon-based labeling dataset is better than the score-based labeling dataset.

---

[6] https://colab.research.google.com.
[7] https://pytorch.org.
[8] https://huggingface.co/transformers.

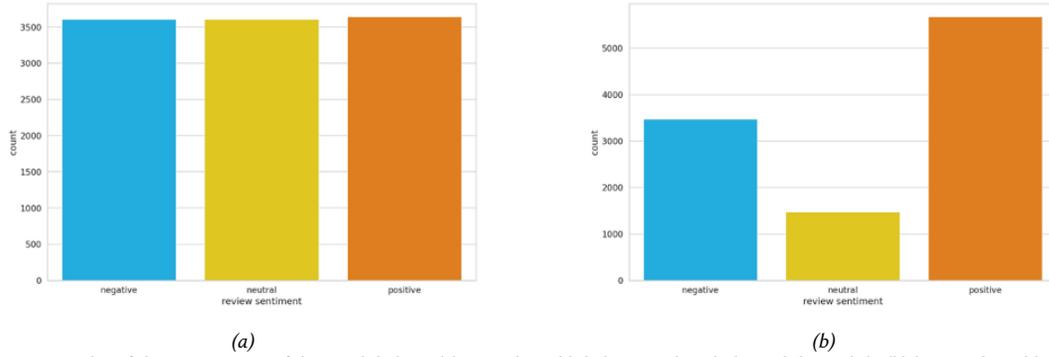

*(a)* *(b)*

Figure 4: Results of the comparison of dataset labeling, (a) score-based labeling produce balanced data while (b) lexicon-based labeling produce imbalanced data

Table 2: Fine-tuning result of pre-trained models for score-based labeling data with learning rate 1e-5 on 10 epochs.

| ML Baseline Model | Score-Based Labeling | | | Lexicon-Based Labeling | | |
|---|---|---|---|---|---|---|
| | Training Accuracy | Cross Validation Accuracy | Test Accuracy | Training Accuracy | Cross Validation Accuracy | Test Accuracy |
| kNN | 0.5738 | 0.35 | 0.3692 | 0.2565 | 0.23 | 0.2504 |
| SVM | 0.9361 | 0.55 | 0.5736 | 0.9901 | 0.72 | 0.7796 |
| Naïve Bayes | 0.8441 | 0.53 | 0.5478 | 0.7281 | 0.63 | 0.6629 |
| Decision Tree | 0.9545 | 0.50 | 0.5027 | 1.0 | 0.69 | 0.7241 |
| Random Forest | 0.9545 | 0.54 | 0.5681 | 1.0 | 0.76 | 0.7890 |

Table 3: Fine-tuning result of pre-trained models for score-based labeling data with learning rate 1e-5 on 10 epochs.

| Pre-trained | Batch Size | Avg Training Accuracy | Avg Validation Accuracy | Training Time | Test Accuracy |
|---|---|---|---|---|---|
| BERT-Base multilingual | 16 | 0.6993 | 0.5832 | 13min 29s | 0.5616 |
| BERT-Base multilingual | 32 | 0.6996 | 0.5882 | 10min 04s | 0.5598 |
| IndoBERT-Base | 16 | 0.7982 | 0.6093 | 12min 58s | 0.5856 |
| IndoBERT-Base | 32 | 0.7907 | 0.5911 | 09min 45s | 0.5598 |

Table 4: Fine-tuning result of pre-trained models for lexicon-based labeling data with learning rate 1e-5 on 10 epochs.

| Pre-trained | Batch Size | Avg Training Accuracy | Avg Validation Accuracy | Training Time | Test Accuracy |
|---|---|---|---|---|---|
| BERT-Base multilingual | 16 | 0.6482 | 0.6726 | 12min 49s | 0.7231 |
| BERT-Base multilingual | 32 | 0.8697 | 0.7881 | 09min 50s | 0.8003 |
| IndoBERT-Base | 16 | 0.9329 | 0.8344 | 15min 41s | 0.8248 |
| IndoBERT-Base | 32 | 0.9337 | 0.8218 | 10min 59s | 0.8229 |

The first fine-tuning experiment for two pre-trained models in score-based labeling data performed using base learning rates [1e-5] at 16 and 32 batch sizes with ten epochs are shown in Table 3. The first fine-tuning of BERT for two pre-trained models in score-based labeling data performed using base learning rates 1e-5 at 16 and 32 batch sizes with ten epochs are shown in Table 4. IndoBERT-Base with a batch size of 16 shows the best accuracy in testing accuracy of 0.5856. However, the results obtained show that the fine-tuning of the BERT model on the dataset with score-based labeling is not much better than the SVM of 0.5736 at the ML baseline. In the second fine-tuning experiment, we use the same hyperparameter with previous for lexicon-based labeling data, the result is shown in Table 4. IndoBERT-Base still obtained the best test accuracy results at batch size 16 of 0.8248. This is because BERT-Base multilingual is only trained on Wikipedia corpus, even though the dataset used contains a lot of slang sentences. While indoBERT-Base is trained in the larger data in Indonesian and contains formal and slang language, such as Twitter. In the last experiment, we only focused on lexicon-based labeling data because the dataset with the score-based labeling showed unfavorable results. We only use IndoBERT-Base because based on previous tests, it shows stable results. We still used learning rate variations in the [1e-5, 2e-5, 3e-5] at batch sizes 32, but we increased the epochs to 25. We wanted to increase the accuracy obtained in the previous test. The best test accuracy is obtained when the learning rate is 2e-5 in batch size 16 with a value of 0.8493, as shown in Table 5. The best fine-tuning history of last experiments based on Table 5 is shown in Figure 5. As a result of all experiments, a smaller batch size increases the overall computation time. IndoBERT-Base as a pre-trained model that is specific trained in Indonesian shows the best results than the pre-trained multilingual.

Table 5: IndoBERT-Base fine-tuning results for lexicon-based data labeling on 25 epochs

| Batch Size | Learning Rate | Avg Training Accuracy | Avg Validation Accuracy | Training Time | Test Accuracy |
|---|---|---|---|---|---|
| 16 | 1e-5 | 0.9701 | 0.8215 | 23min 21s | 0.8267 |
| 32 | 1e-5 | 0.9709 | 0.8301 | 16min 40s | 0.8342 |
| 16 | 2e-5 | 0.9650 | 0.8337 | 24min 20s | 0.8493 |
| 32 | 2e-5 | 0.9718 | 0.8413 | 16min 56s | 0.8286 |
| 16 | 3e-5 | 0.9332 | 0.8141 | 24min 52s | 0.7909 |
| 32 | 3e-5 | 0.9678 | 0.8269 | 16min 55s | 0.8173 |

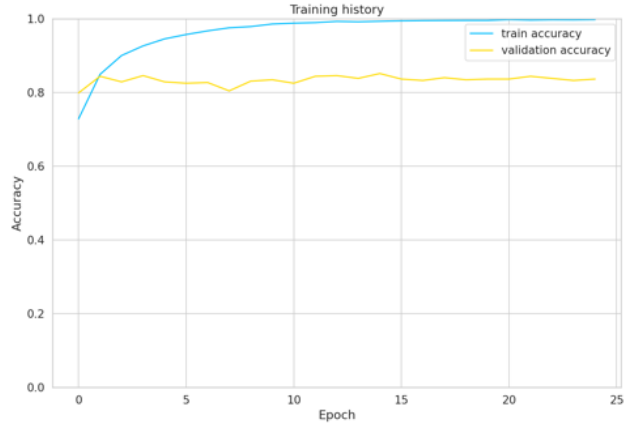

Figure 5: Training and validation accuracy on the best fine-tuning result. IndoBERT-Base with learning rate=2e-5; batch size=16; number of epochs=25

## 4 CONCLUSION

This study has successfully conducted in-depth experiments on analyst sentiment on Indonesian language user reviews. BERT being the newest state-of-the-art natural language processing has shown optimal results for downstream tasks, such as sentiment analysis. Through fine-tuning, pre-trained models can do transfer learning. So, it is much better than machine learning. The method of labeling the data affects the final result. Lexicon-based labeling is proven to show much better accuracy than score-based labeling. The best results are obtained with the IndoBERT-Base model. The highest accuracy is 84%, with 25 epochs and a training time of 24 minutes. Future work is to investigate the specific effect of each preprocessing procedure and other settings related to the tuning to further language model for sentiment analysis purposes.